\documentclass[letterpaper, 10 pt, conference]{ieeeconf}  
 \pdfoutput=1

\IEEEoverridecommandlockouts                              

\usepackage{cite}
\usepackage{graphicx}
\graphicspath{{./figures/}}

\usepackage{tikz}
\usetikzlibrary{fit}					
\usetikzlibrary{backgrounds}	
\usetikzlibrary{arrows}
\usepackage{multirow}
\usepackage{cellspace}
\usepackage{multicol} 

\usepackage[caption=false]{subfig}
\newcommand{\Definition}{\textit{Definition:}}
\usepackage{xcolor}
\usepackage{tikz}
\usetikzlibrary{decorations.markings}
\usetikzlibrary{shapes.geometric}

\pgfdeclarelayer{edgelayer}
\pgfdeclarelayer{nodelayer}
\pgfsetlayers{edgelayer,nodelayer,main}

\tikzstyle{none}=[inner sep=0pt]
\definecolor{hexcolor0xff0000}{rgb}{1.000,0.000,0.000}
\definecolor{hexcolor0x000000}{rgb}{0.000,0.000,0.000}
\definecolor{hexcolor0x00ff00}{rgb}{0.000,1.000,0.000}
\definecolor{hexcolor0x000000}{rgb}{0.000,0.000,0.000}
\definecolor{hexcolor0xff0000}{rgb}{1.000,0.000,0.000}
\definecolor{hexcolor0xffff00}{rgb}{1.000,1.000,0.000}
\definecolor{hexcolor0x000000}{rgb}{0.000,0.000,0.000}
\definecolor{hexcolor0xffffff}{rgb}{1.000,1.000,1.000}
\definecolor{hexcolor0x000000}{rgb}{0.000,0.000,0.000}
\definecolor{hexcolor0x000000}{rgb}{0.000,0.000,0.000}

\tikzstyle{rn}=[circle,fill=hexcolor0xff0000,draw=hexcolor0x000000,line width=0.8 pt]
\tikzstyle{gn}=[circle,fill=hexcolor0x00ff00,draw=hexcolor0x000000,line width=0.8 pt]
\tikzstyle{graph}=[rectangle,text width=70pt,  rounded corners, text centered, fill=hexcolor0xff0000,draw=hexcolor0x000000,line width=0.8 pt]
\tikzstyle{robot}=[rectangle, text width=70pt, text centered, rounded corners, text height= 10pt, fill=hexcolor0xffff00,draw=hexcolor0x000000, minimum width=60pt]

\tikzstyle{oneway}=[-latex,draw=hexcolor0x000000, line width=2.00]
\tikzstyle{arrow}=[-,draw=hexcolor0x000000,postaction={decorate},decoration={markings,mark=at position .5 with {\arrow{>}}},line width=2.000]
\tikzstyle{tick}=[-,draw=hexcolor0x000000,postaction={decorate},decoration={markings,mark=at position .5 with {\draw (0,-0.1) -- (0,0.1);}},line width=2.000]
\tikzstyle{bothways}=[latex-latex,draw=hexcolor0x000000,line width=3.000]

\begin{document}
%
\title{Model-free control framework for multi-limb soft robots}
%
%
%

\author{Vishesh~Vikas,~\IEEEmembership{Member,~IEEE,}
        Piyush~Grover,
        and~Barry~Trimmer,~\IEEEmembership{Member,~IEEE}
\thanks{V. Vikas and B. Trimmer are with the Neuromechanics and Biomimetics Lab, Tufts University, MA 02155 USA e-mail: \{vishesh.vikas , barry.trimmer \}@tufts.edu}
\thanks{P. Grover is with Mitsubishi Electric Research Laboratories, Cambridge MA 02139 USA e-mail: grover@merl.com}}

%
%

\markboth{Model-free control framework for multi-limb soft robots}%
{Model-free control framework for multi-limb soft robots}
%



\maketitle

\begin{abstract}
The deformable and continuum nature of soft robots promises versatility and adaptability. However, control of modular, multi-limbed soft robots for terrestrial locomotion is challenging due to the complex robot structure, actuator mechanics and robot-environment interaction. Traditionally, soft robot control is performed by modeling kinematics using exact geometric equations and finite element analysis. 

The research presents an alternative, model-free, data-driven, reinforcement learning inspired approach, for controlling multi-limbed soft material robots. This control approach can be summarized as a four-step process of discretization, visualization, learning and optimization. The first step involves identification and subsequent discretization of key factors that dominate robot-environment, in turn, the robot control. Graph theory is used to visualize relationships and transitions between the discretized states. The graph representation facilitates mathematical definition of periodic control patterns (simple cycles) and locomotion gaits. Rewards corresponding to individual arcs of the graph are weighted displacement and orientation change for robot state-to-state transitions. These rewards are specific to surface of locomotion and are learned. Finally, the control patterns result from optimization of reward dependent locomotion task (e.g. translation) cost function. The optimization problem is an Integer Linear Programming problem which can be quickly solved using standard solvers.

The framework is generic and independent of type of actuator, soft material properties or the type of friction mechanism, as the control exists in the robot's task space. Furthermore, the data-driven nature of the framework imparts adaptability to the framework toward different locomotion surfaces by re-learning rewards.
\end{abstract}


%
\IEEEpeerreviewmaketitle

\section{Introduction}
Roboticists in recent years have been inspired by the ability of animals to leverage structural soft materials for locomotion and manipulation tasks. This has resulted in the development of soft material robots powered by a variety of actuators including bio-inspired soft \cite{laschi_design_2009, cianchetti_new_2009} and continuum manipulators \cite{walker_continuum_2005}, rigid link-based snake-like robots \cite{hirose_biologically_2004, wright_design_2007}, pneumatic soft multi-gait robots \cite{shepherd_multigait_2011} and shape memory alloy actuated soft robots \cite{lin_goqbot:_2011,  umedachi_highly_2013, wang_locomotion_2014}. The control of flexible-link robots has been done using model-based or model-free control approaches \cite{rigatos_model-based_2009}. The soft continuum manipulator control is traditionally performed using continuum modeling techniques \cite{hannan_kinematics_2003, webster_design_2010, renda_3d_2012}, while non-continuous curvature soft robots have been controlled using fast finite element methods \cite{duriez_control_2013}. Soft robots capable of terrestrial locomotion interact extensively with the environment and manipulate friction to facilitate movement. The control of such robots using model-based approach will involve detailed mathematical descriptions \cite{chirikjian_kinematics_1995} of the robot kinematics, dynamics, actuator mechanics and, most importantly, the robot-environment interaction. This approach is computationally intensive \cite{saunders_modeling_2011} and robot specific. In this research, we present an alternative model-free approach that is generic and adaptable.

The following sections describe the mathematical basis for model free control as the representation of state transitions on a directed graph then solving the gait optimization problem. An experimental test case of three limbed soft robot is presented, followed by a discussion of the general utility of the approach.%

\textit{Contribution:} %
The research presents a generic, adaptive, data-driven model-free control framework for locomotion control of multi-limbed terrestrial soft robots. The framework can be summarized as a four-step process of 1) \textit{discretizing} key factors dominating robot control via robot-environment interaction, 2) using graph theory to \textit{visualize} relationships between discretized robot states, 3) \textit{learning} the surface-dependent results of state transitions and 4) \textit{optimizing} desired cost function to obtain a control sequence. Computationally, the optimization problem is an Integer Linear Programming (ILP) problem that can be solved using standard solvers. The use of graph theory facilitates mathematical definition of periodic control patterns (simple cycles) that form linear basis to locomotion gaits. Furthermore, the graph representation introduces robustness into the control framework by making it fault-tolerant.
\section{Model-free control framework}
Controlling locomotion by soft robots is challenging due to the difficulty to accurately model most robot-environment interactions. This interaction is easier to model in fluids because there are good mathematical tools for describing force propagation in continuous media, however, modeling discontinuous terrestrial interactions is much more complex. Furthermore, the modeling of the soft robot may be restricted by shape (for analytical continuum solutions)  \cite{chirikjian_kinematics_1995, webster_design_2010} or involve simplification/discretization \cite{saunders_modeling_2011}. The complexity is further increased by actuator-specific modeling. As an example, the properties of shape memory alloy actuators (SMAs) change over time because heat flux cannot be controlled very precisely in natural settings.
The model-free control approach takes inspiration from reinforcement learning \cite{sutton_introduction_1998} by focusing on goal-directed learning. This approach does not directly model the robot kinematics, the actuator or the robot-environment interaction, but indirectly accounts for the robot-environment interaction by observing the effects of changes in the robot-environment interactions. In the related model reduction literature, this approach is often called the input-output approach\cite{antoulas_approximation_2005}. Usually the input-output approach is preferred when the full dynamics of the system are complicated, and the input action is relatively limited.

In this work, the robot locomotion is formulated as a class of optimization problems on directed graphs. An analogy between the language of graph theory and soft robot locomotion is constructed. This analogy is constructed alongside an example of a robot to facilitate better understanding of the framework.

\begin{figure}
\centering
\includegraphics[width=0.75\columnwidth]{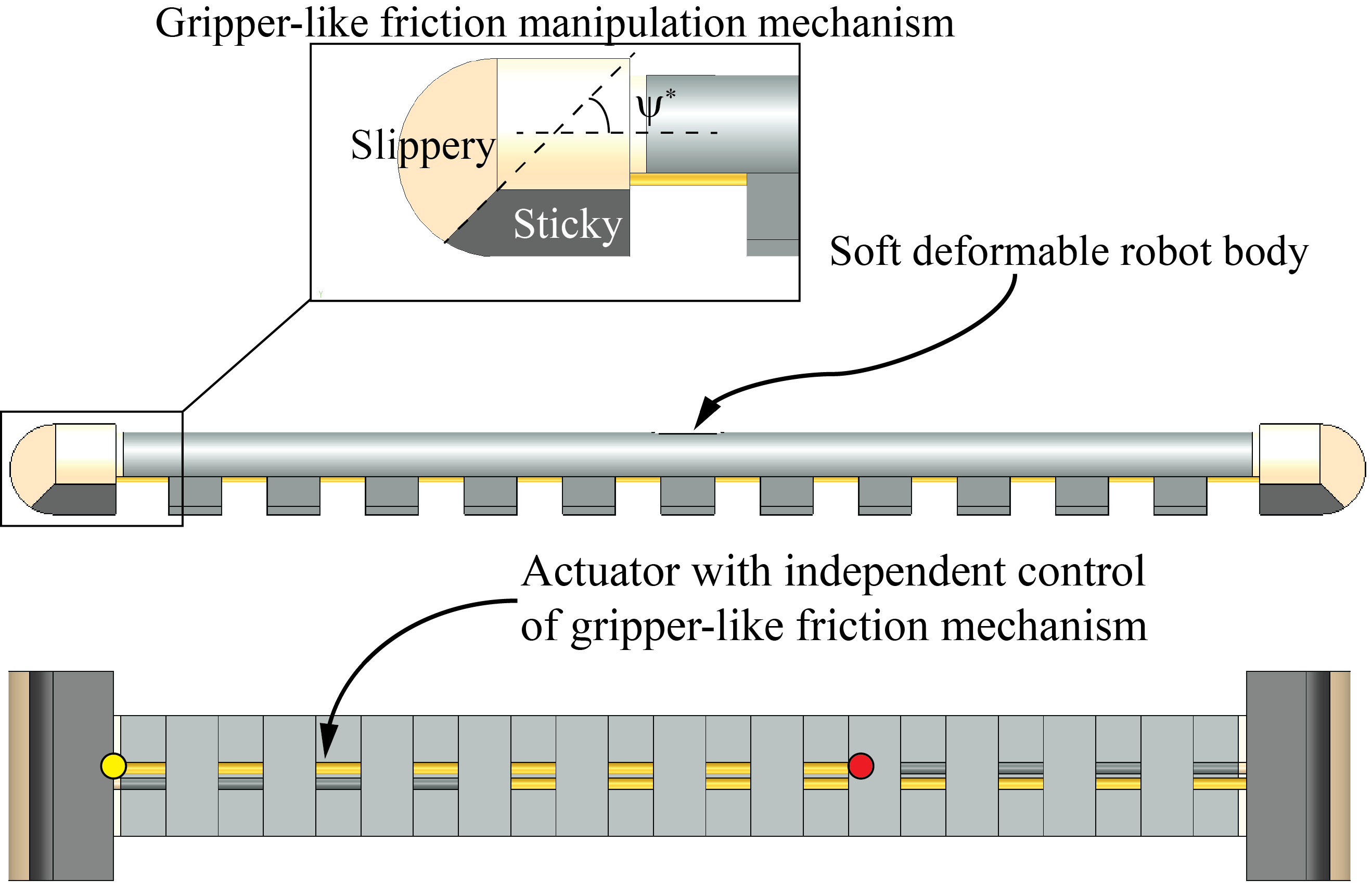}
\caption{Example soft robot comprises of soft, deformable body and gripper-like friction manipulation mechanisms at both ends (front and rear) of the robot. The actuators (gold channel starting from yellow circle and ending at red circle) independently control each friction manipulation mechanism.}
\label{Fig:ExampleRobot}
\end{figure}

\textit{Example robot:} The example robot is a monolithic 3-D printed soft robot with a soft deformable body and two gripper-like (friction manipulation) mechanisms at each end of the robot as illustrated in Figure \ref{Fig:ExampleRobot}.

\subsection{Discretization of robot-environment interaction}
Locomotion results from manipulation and optimization of friction forces at different parts of a body \cite{radhakrishnan_locomotion:_1998} . This may be performed using directional friction or with a mechanical or chemical mechanism \cite{gorb_attachment_2001, hirose_biologically_2004}. The body-environment interaction can be discretized into small number of finite behaviors.

\Definition~\textit{Behavior}, denoted by $B$, are discrete behaviors of a system part e.g. for a robot sub-system - grip on/off ($B_{G}$), directional friction ($B_{DF}$). 
\begin{eqnarray}
\label{Eqn:Grip}
B_{G} &=& \left\{ \begin{array}{cc}
0 & \mathrm{for\ grip\ on}\\
1 & \mathrm{for\ grip\ off}
\end{array}
\right.\\
\label{Eqn:DirectionalFriction}
B_{DF} &=& \left\{ \begin{array}{cc}
0 & \mathrm{movement\ in\ preferred\ direction}\\
1 & \mathrm{movement\ in\ opposite\ direction}
\end{array}
\right.
\end{eqnarray}
Typically, an actuator independently controls the behavior of a robot sub-system, but, this behavior representation of robot-environment interaction is independent of the \textit{type} of actuator.

\Definition~\textit{State}, denoted by $S$, consists of the corresponding behaviors of all robot sub-systems. The total number of states $n$ for a robot with $m$ sub-system robot parts, each having $b$ discrete behaviors can be defined as 
\begin{equation} \label{Eqn:Nodes}
n = b^m
\end{equation}
The example soft robot has two gripper-like friction manipulation mechanisms ($m=2$ sub-systems) that have binary behaviors ($b=2$) as evident from Equation \ref{Eqn:Grip}. As a result, the total number of states are $n=4$ and can be exhaustively written as $\{(00),(01),(10),(11)\}$.
\subsection{Visualization, Learning and Graph Theory}
\Definition~Each graph \textit{node} represents one robot state. The nodes are denoted by $\mathbf{N_i}$ for $i=1,\cdots n$.  

\begin{figure}
\centering
\begin{tikzpicture}[->,>=stealth',shorten >=1pt,auto,node distance=3cm,
  thick,%
  scale=0.9, every node/.style={scale=0.9},%
  main node/.style={circle,fill=blue!20,draw,font=\sffamily\Large\bfseries}]

  \node[main node] (1) {$\mathbf{N_1}$};
  \node[main node] (2) [below left of=1] {$\mathbf{N_2}$};
  \node[main node] (3) [below right of=2] {$\mathbf{N_3}$};
  \node[main node] (4) [below right of=1] {$\mathbf{N_4}$};

  \path[every node/.style={font=\sffamily\small}]
    (1) edge [bend right] node[left] {$A_1$} (2)
    	edge [bend right] node [left] {$A_3$} (3)
    	edge [bend left] node [left] {$A_5$} (4)    
    (2) edge  node [left] {$A_2$} (1)
        edge [bend right] node[left] {$A_7$} (3)
        edge [bend right] node {$A_9$} (4)
    (3) edge [bend right] node [right] {$A_4$} (1)
        edge node[right] {$A_8$} (2)
        edge [bend right] node[right] {$A_{11}$} (4)
    (4) edge node [left] {$A_6$} (1)
        edge [bend right] node  {$A_{10}$} (2)
        edge node [right] {$A_{12}$} (3);
\end{tikzpicture}
\caption{The directed graph corresponding to the example robot comprising of two system parts ($m=2$) such that each part has two discretized behaviors ($P=2$). The $n=4$ nodes are $\mathbf{N_1,\ N_2,\ N_3,\ N_4}$. They correspond to the four robot states $\{00\}, \{01\}, \{10\}, \{11\}$. }
\label{Fig:ExampleGraph}
\end{figure}
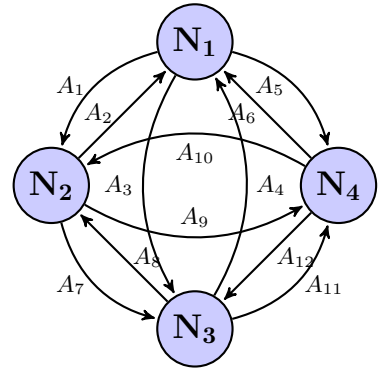

Each graph\cite{diestel_graph_2010}  \textit{directed arc} is a connection between two different nodes. The total number of arcs for the case where the robot can transition from any node to another (fully connected graph) are $P=n\cdot (n-1)$. Each arc is identified as $A_k$ for $k=1,2,\cdots,P$. These nodes connected by directed arcs comprise of a directed graph. The directed graph for the example soft robot with $P=12$ arcs is shown in Figure \ref{Fig:ExampleGraph}.  

The arcs represent the transition from one robot state to another. This \textit{robot state transition}, equivalent of change in robot-environment interaction, will result in some translation ($\Delta x, \Delta y$) and rotation ($\Delta \theta$) on a plane. The weighted result is called the \textit{state transition reward}. The state transition reward vector $R_i \in \mathcal{R}^{3\times 1}$ corresponding to vector arc weight of arc $A_i$ is written as
\begin{eqnarray} \label{Eqn:RewardVector}
R_i &=& \left[w_{ix}, w_{iy}, w_{i \theta}\right]^T \qquad for\ i = 1, \cdots P
\end{eqnarray}

The full reward matrix for the graph is 
\begin{eqnarray} \label{Eqn:RewardMatrix}
\mathrm{R} &=& \left[ R_1\: R_2\:  \cdots \:R_P \right], \qquad \mathrm{R}\in \mathcal{R}^{3 \times P}
\end{eqnarray}
\textit{Learning.} The state transition rewards are experimentally determined and change with the surface of contact as they attempt to indirectly model the robot-environment interaction. This learning ability imparts adaptability to the framework by facilitating compensation (learning) for unexpected changes in the environment.

\Definition~\textit{Simple cycle}. A closed walk consists of a sequence of nodes starting and ending at the same node, with each two consecutive nodes in the sequence connected by a directed arc. A simple cycle is a closed walk with no repetitions of nodes and directed arcs allowed, other than the repetition of the starting and the ending node.  Simple cycles may also be described by their sets of directed arcs, unlike closed walks for which the multi-set of arcs does not unambiguously determine the node ordering. A simple cycle $c_i \in \mathcal{R}^P$ for $i=1,\cdots, K$ where $K$ is the total number of cycles and
\begin{equation} \label{Eqn:SimpleCyleVector}
c_{i,j} = \left\{\begin{array}{cc}
1 & \mathrm{if\ }c_i\ \mathrm{includes\ arc\ }A_j\\
0 & otherwise
\end{array}
\right.
\end{equation}

Given the graph structure (i.e. set of nodes, and directed arcs), the problem of finding all simple cycles can be solved efficiently\cite{johnson_finding_1975}. We use the open source software NetworkX\cite{hagberg_aric_networkx._????} to obtain the solution of this problem. While the graph is fully connected, many of the state transitions do not lead to any significant locomotion. Hence, the directed arcs with all rewards below a certain threshold can be removed from the graph structure before doing the gait computation.

The simple cycles are periodic cycles of state transitions and act as linear basis for finding locomotion gaits (circulation). The reward vector associated with every individual simple cycle is referred to as the simple cycle reward $J_{i} \in \mathcal{R}^3$
\begin{equation}
J_i = Rc_i
\end{equation}

For the example soft robot graph structure, a simple cycle $c_{eg}$ of $\{N_1 \rightarrow N_2 \rightarrow N_3 \rightarrow N_4 \rightarrow N_1\}$ will comprise of arcs $A_1, A_7, A_{11}, A_6$ corresponding to the numbering given in Figure \ref{Fig:ExampleGraph}. Hence, the $c_{eg} \in \mathcal{R}^{12}$ is written as
\begin{equation}
c_{eg,j} = \left\{ \begin{array}{cc}
1 & \mathrm{for}\ j=1,6,7,11\\
0 & \mathrm{otherwise}
\end{array}
\right.
\end{equation}

\Definition~A \textit{circulation} is a linear integer combination of simple cycles 
\begin{equation} \label{Eqn:Circulation}
L = \sum\limits_{i=1}^{K} x_i c_i,\qquad x_i \in \{0,1,2,..\}
\end{equation}
We define \textit{locomotion gait} as being equivalent to a circulation. It is important to note that this notation does not define the order of simple cycles, rather, only the combination. The reward for the circulation can be similarly written as
\begin{equation} \label{Eqn:CirculationReward}
J(L) = \sum\limits_{i=1}^{K} x_i J_{i},\qquad x_i \in \{0,1,2,..\}
\end{equation}
The non-ambiguous representation of locomotion gaits using simple cycles, which form the linear basis, is very important as it allows for a simple formulation of the optimization problem.
%
\subsection{Optimization and Finding Gaits} \label{opt}
The locomotion gaits of robots optimize some cost function e.g. maximize translation or rotation for a given number of steps. For such analysis, the reward for a circulation $L$ is decomposed into three components, corresponding to $x,y,\theta$ components
\begin{equation}
J(L)=[J^x,J^y,J^\theta]^T
\end{equation}
Consider the problem of finding a locomotion gait which maximizes the translation in $+X$ direction. We impose a constraint on the maximum number of state transitions allowed $\left( len(L) = \sum\limits_{i=1}^{K}\sum\limits_{j=1}^{p}x_ic_{i,j} \right)$ and permitted residual translation in $+Y$ direction ($J_y$) and rotation in $+\theta$ direction ($J_\theta$). 
Hence, the optimization problem is written as
\begin{equation} \label{Eqn:OptimizationCost}
\max\limits_{x\in \left(\mathcal{Z}^+\right)^P} J^x
\end{equation}
with constraints
\begin{equation}\label{Eqn:OptimizationConstraints}
J^y \in [-\epsilon_{y-}, \epsilon_{y+}],\ %
J^\theta \in [-\epsilon_{\theta-}, \epsilon_{\theta+}],\ %
len(L) \leq l_{max}
\end{equation}
This problem is an integer linear programming (ILP) problem and can be solved for small to medium size graphs using one of the many standard solvers such as Matlab optimization toolbox \cite{lopez_cesar_optimization_????} and Gurobi\cite{optimization_gurobi_and_others_gurobi_????}.

\begin{figure}
\centering
\includegraphics[width=0.75\columnwidth]{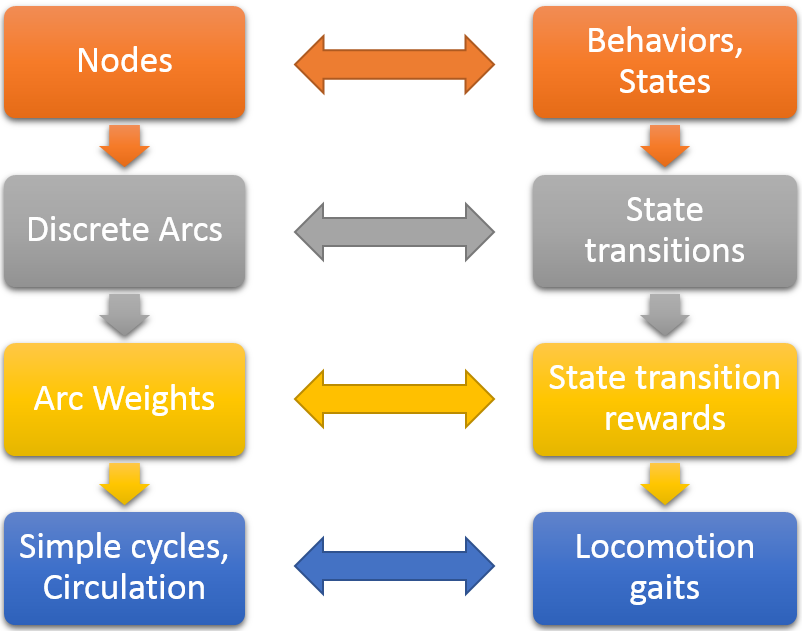}
\caption{Analogy between directed graph and robot mechanics for the model-free control framework}
\label{Fig:Analogy}
\end{figure}

This analogy of soft robot locomotion with graph theory (Figure \ref{Fig:Analogy}) is very advantageous - 1) The graph representation of discretized robot behaviors facilitates easy visualization, 2) the unique mathematical representation of simple cycle vector (Equation \ref{Eqn:SimpleCyleVector}) is instrumental in defining the 3) circulation-locomotion gait analogy (Equation \ref{Eqn:CirculationReward}). 4) The optimization problem is an ILP problem (Equations \ref{Eqn:OptimizationCost}, \ref{Eqn:OptimizationConstraints}) which can be quickly solved using standard linear solvers for small to medium size graphs.

\textit{Speed.} The periodic control sequences (simple cycles) are independent of actuation and material variations. Hence, the speed of locomotion is solely dependent on speed at which a robot can transition from one state to another. As an example, given two same soft robots $R_1,\ R_2$ actuated by actuators (e.g. motors) $M_1,\ M_2$ with power $P_1,\ P_2$ such that $P_2>P_1$. Here, the actuator $M_2$ facilitates faster transition from one state to another, therefore, $R_2$ is capable of faster locomotion than $R_1$ with the same control sequence. Similar argument also holds for robots designed using two different materials one having faster rate of deformation than the other.

\subsection{Extensions to other locomotion tasks}
The framework described in section \ref{opt} can be generalized to include more complicated gaits and locomotion objectives. In our formulation so far, we have assumed that the reward matrices remain effectively independent of the robot coordinate system. This assumption holds only approximately for the case when $\theta$ is small, but in case of larger $\theta$ displacement, the displacement in $x$ and $y$ direction is multiplied by a rotation matrix. The resulting locomotion optimization problem essentially becomes a nonlinear integer programming problem. The resulting gaits can have arbitrarily large angular displacement, and can lead to highly complex gaits. Furthermore, this framework could also be adapted to have the robot follow a given curved path. This extension will be the subject of future work.

The robustness and fault-tolerance of the framework is portrayed by its ability to efficiently respond to scenarios such as the loss of a limb. In this case, one of the actuators/sub-systems of the robot becomes inoperable. A runtime modification of the graph structure (by removing the corresponding nodes and arcs), followed by re-computation of the optimal gaits can handle this situation.

Furthermore, given the data on a certain class of actuators, the length constraint of maximum number of state transitions in equation \ref{Eqn:OptimizationConstraints} can be replaced by a weighted constraint i.e. $t(L)=\sum\limits_{i=1}^K\sum\limits_{j=1}^px_iT_jc_{i,j} < t_{max}$, where $T\in \mathcal{R}^p$ is vector containing time taken for each state transition, and $t_{max}$ is the maximum allowable gait time. 

\section{Experiment}
\begin{figure}
\centering
\includegraphics[width=0.75\columnwidth]{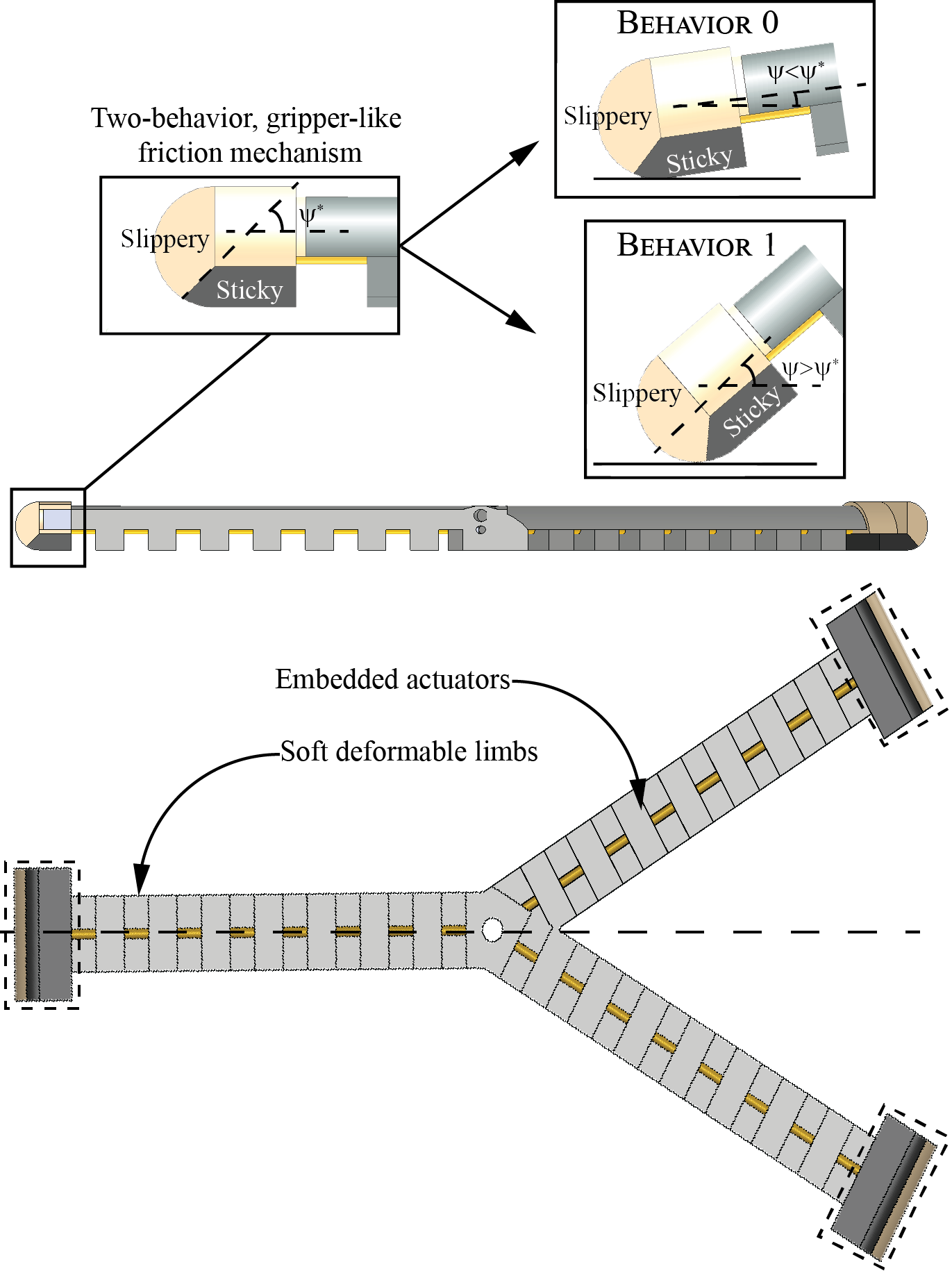}
\caption{Three limb robot with deformable soft body and a gripper-like friction mechanism at end of each limb. The gripper-like friction mechanism is made using two materials with different coefficients of friction - a sticky/soft and slippery/hard material. The material of contact changes with the shape of the robot and can exist in two discrete behaviors - $0$ and $1$ such that the switch happens about the critical contact angle $\psi^*$. The behavior of these friction mechanisms is independently controlled (via the limb shape) using three embedded shape memory alloy (SMA) actuators.}
\label{Fig:YbotDefinition}
\end{figure}

The experimental soft robot is similar to the example soft robot, but has three limbs with gripper-like two state friction mechanisms at end of each limb as shown in Figure \ref{Fig:YbotDefinition}. The soft robot has a soft deformable body made of rubber-like TangloPlus\texttrademark~and is printed on Connex 500\texttrademark~multi-material 3D printer. Each of the friction mechanism uses soft rubber-like TangoPlus\texttrademark as the sticky material and hard abs-like VeroClear\texttrademark~as the slippery material. 

These gripper-like mechanisms are independently controlled using NiTi SMA actuators (Toki Corporation \textregistered ). The SMA  coils  are  electrically activated to shorten by joule heating and they relax to the original shape using the stored elastic energy upon deactivation. SMA activation change both the limb shape and its friction state as the contact angle between the limb and the surface varies about the critical contact angle $\psi^*$ (Figure \ref{Fig:YbotDefinition}). The properties of SMA coils may vary over time due to inconsistent cooling, etc. but the discretization of sub-system behavior (friction mechanism) makes the control sequence independent of the precise condition of the actuator, and merely dependent on the contact angle. This soft robot has three ($m=3$) sub-systems each having two discrete behaviors ($b=2$), thus, the total number of states are $n=2^3$ (visual representation - Appendix \ref{App:Visualization}) such that the node $\mathbf{N_i}$ and robot state analogy can be expressed as
\begin{equation}
\mathbf{N_i} \Leftrightarrow \mathrm{dec2bin}(i-1) \qquad i = 1,2..., 8
\end{equation}
where $\mathrm{dec2bin}$ function converts decimal to binary format. This fully connected directed graph comprises of $P=8\cdot(8-1) = 56$ arcs. The experiment is run on a smooth planar surface (table-top) and the state transition rewards (arc weights) are recorded as weighted mean of the relative change in position and orientation for 10 state transition repetitions (Appendix \ref{App:Rewards}). The weighted state transition rewards are dependent on the robot and the surface of contact.

The solution to the optimization problem for a given maximum length ($l_{max}$) and residual motion ($\epsilon_{\pm y}, \epsilon_{\pm \theta}$)
are integers ($x_i$) corresponding to the number of simple cycles ($c_i$) in the gait. Here, we analyze the different simple cycles obtained from the optimization.

\textit{Translation in $+X$ direction:} Translation in $+X$ is the solution of the optimization problem stated in Equations \ref{Eqn:OptimizationCost}, \ref{Eqn:OptimizationConstraints}. The simple cycle control sequences for $l_{max} = 15$ with tolerances $\epsilon_{\pm y} = 1, \epsilon_{\pm \theta} = 5$ result in the sequences shown in Figure \ref{Fig:X+Translation}. The use of different simple cycles to obtain same goal ($+X$ translation) is an important result as shown in Figure \ref{Fig:X+Translation}.
\newcommand{\thescale}{0.25}
\begin{figure}[h]
\centering
\subfloat[]{
\includegraphics[scale=\thescale]{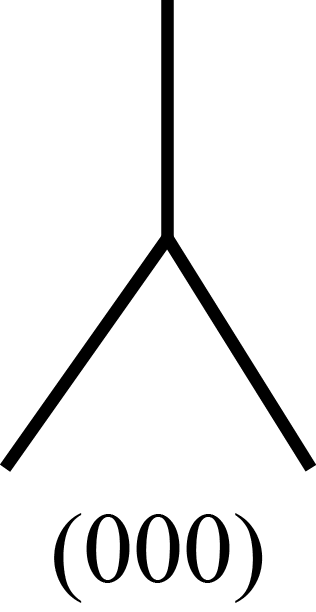} 
\includegraphics[scale=\thescale]{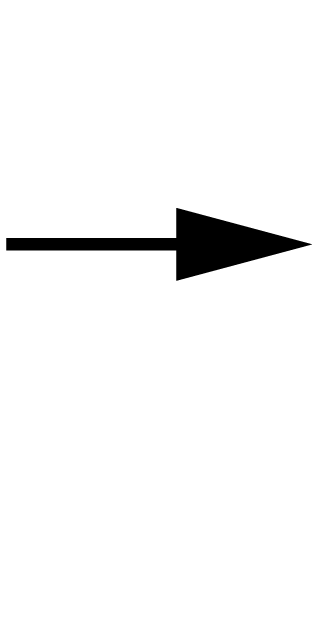}  
\includegraphics[scale=\thescale]{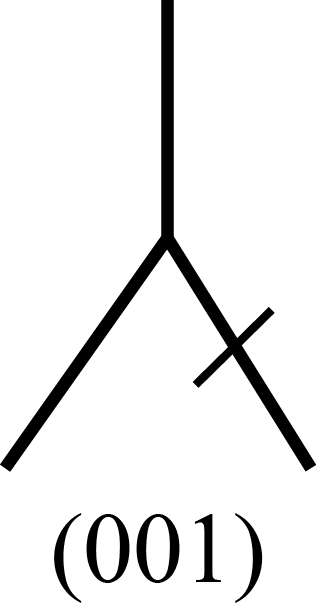} 
\includegraphics[scale=\thescale]{state-arrow-01.png}  
\includegraphics[scale=\thescale]{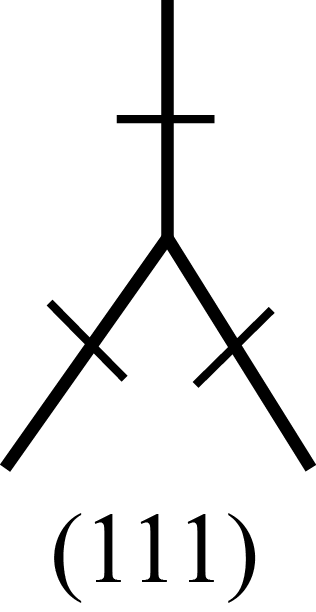} 
\includegraphics[scale=\thescale]{state-arrow-01.png}  
\includegraphics[scale=\thescale]{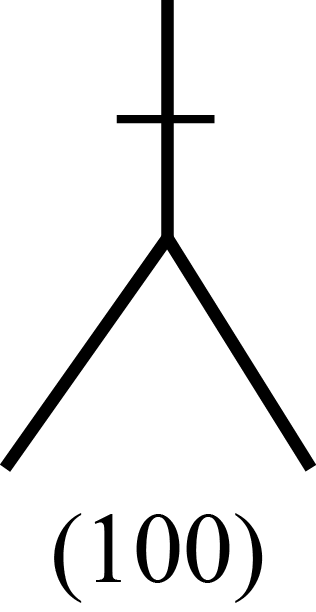} 
\includegraphics[scale=\thescale]{state-arrow-01.png}  
\includegraphics[scale=\thescale]{state-000-01.png} 
}
\newline
\subfloat[]{
\includegraphics[scale=\thescale]{state-000-01.png} 
\includegraphics[scale=\thescale]{state-arrow-01.png}  
\includegraphics[scale=\thescale]{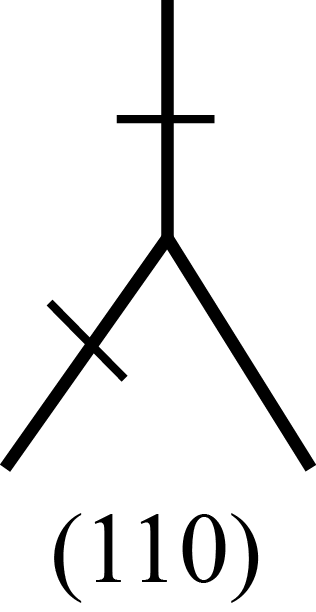} 
\includegraphics[scale=\thescale]{state-arrow-01.png}  
\includegraphics[scale=\thescale]{state-111-01.png} 
\includegraphics[scale=\thescale]{state-arrow-01.png}  
\includegraphics[scale=\thescale]{state-100-01.png} 
\includegraphics[scale=\thescale]{state-arrow-01.png}  
\includegraphics[scale=\thescale]{state-000-01.png} 
}
\newline
\subfloat[]{
\includegraphics[scale=\thescale]{state-111-01.png} 
\includegraphics[scale=\thescale]{state-arrow-01.png}  
\includegraphics[scale=\thescale]{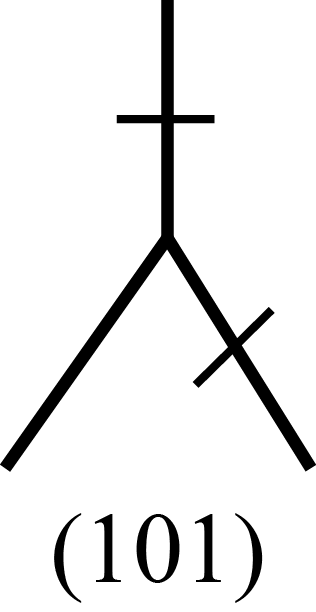} 
\includegraphics[scale=\thescale]{state-arrow-01.png}  
\includegraphics[scale=\thescale]{state-110-01.png} 
\includegraphics[scale=\thescale]{state-arrow-01.png}  
\includegraphics[scale=\thescale]{state-111-01.png} 
}
\caption{Three different state control sequences (simple cycles) resulting from optimization that produce forward translation.}
\label{Fig:X+Translation}
\end{figure}

\begin{figure}[h]
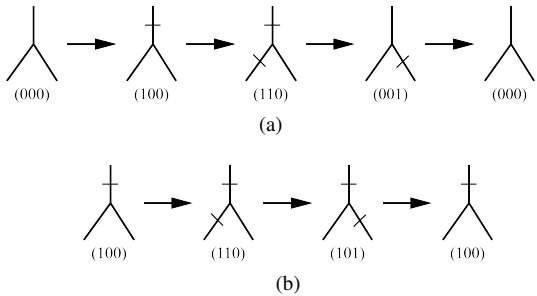

\centering
\subfloat[]{
\includegraphics[scale=\thescale]{state-000-01.png} 
\includegraphics[scale=\thescale]{state-arrow-01.png}  
\includegraphics[scale=\thescale]{state-100-01.png} 
\includegraphics[scale=\thescale]{state-arrow-01.png}  
\includegraphics[scale=\thescale]{state-110-01.png} 
\includegraphics[scale=\thescale]{state-arrow-01.png}  
\includegraphics[scale=\thescale]{state-001-01.png} 
\includegraphics[scale=\thescale]{state-arrow-01.png}  
\includegraphics[scale=\thescale]{state-000-01.png} 
}
\newline
\subfloat[]{
\includegraphics[scale=\thescale]{state-100-01.png} 
\includegraphics[scale=\thescale]{state-arrow-01.png}  
\includegraphics[scale=\thescale]{state-110-01.png} 
\includegraphics[scale=\thescale]{state-arrow-01.png}  
\includegraphics[scale=\thescale]{state-101-01.png} 
\includegraphics[scale=\thescale]{state-arrow-01.png}  
\includegraphics[scale=\thescale]{state-100-01.png} 
}
\caption{Two different simple cycle control sequences resulting from optimization that produce backward translation ($-X$ displacement) of the soft robot.}
\label{Fig:X-Translation}
\end{figure}

\textit{Translation in $-X$ direction:} Calculation of the simple cycle control sequences for optimal $-X$ direction translation can be done by converting the maximization problem to minimization problem. Two simple cycles resulting from modified optimization problem with same constraints are shown in Figure \ref{Fig:X-Translation}.
%

\begin{figure}
\centering
\newcolumntype{D}{>{\centering\arraybackslash} m{80pt} }
\newcolumntype{C}{>{\centering\arraybackslash} m{15pt} }
\cellspacetoplimit 10pt
\cellspacebottomlimit 10pt
\begin{tabular}{SD SC SD}
\includegraphics[height=45pt]{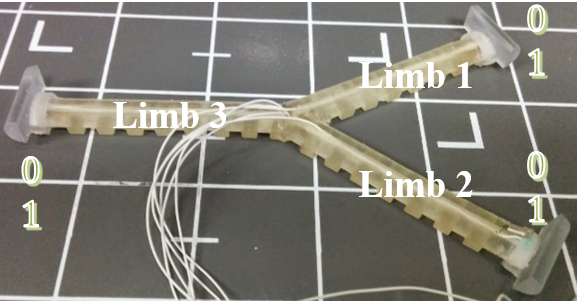} &
\includegraphics[scale=0.15]{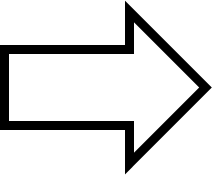}&
\includegraphics[height=45pt]{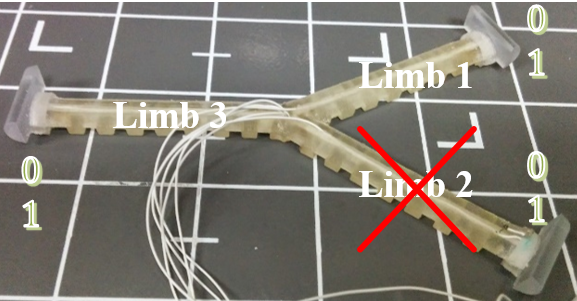} \\
\includegraphics[height=90pt]{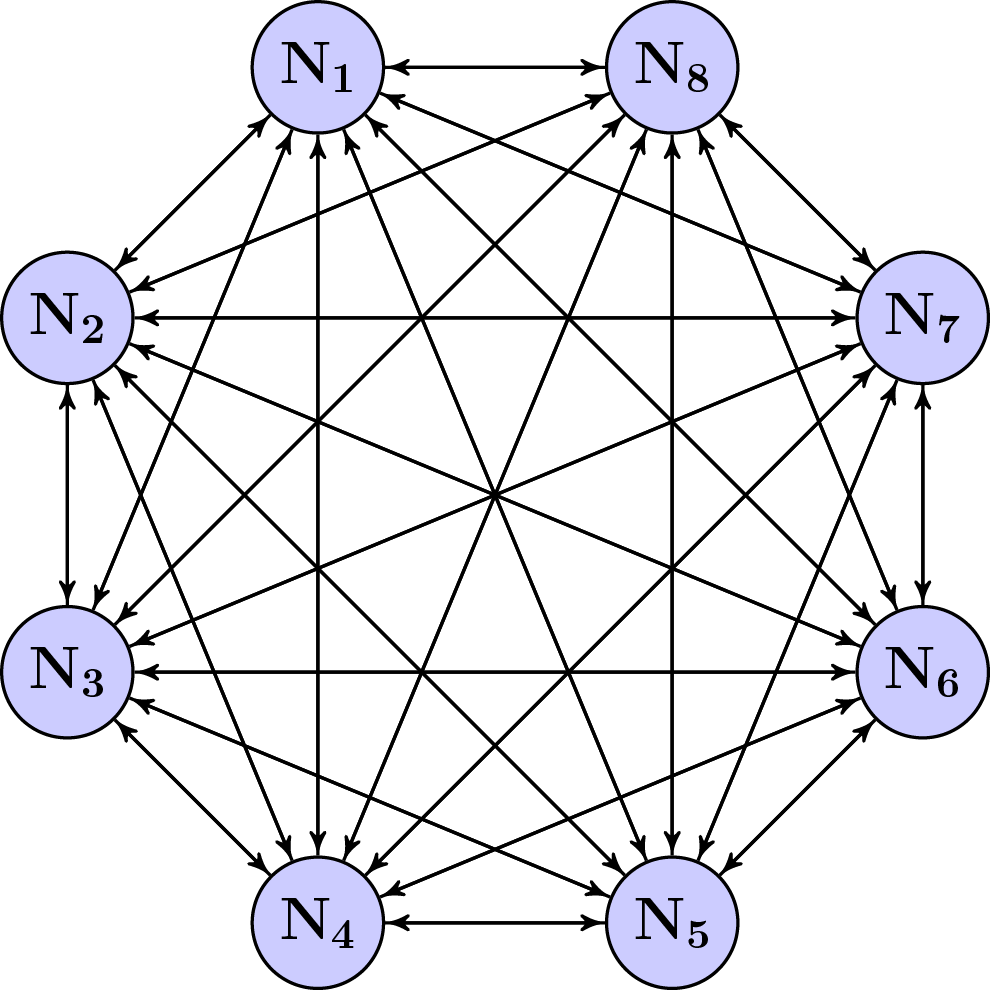}&
\includegraphics[scale=0.15]{arrow}&
\includegraphics[height=90pt]{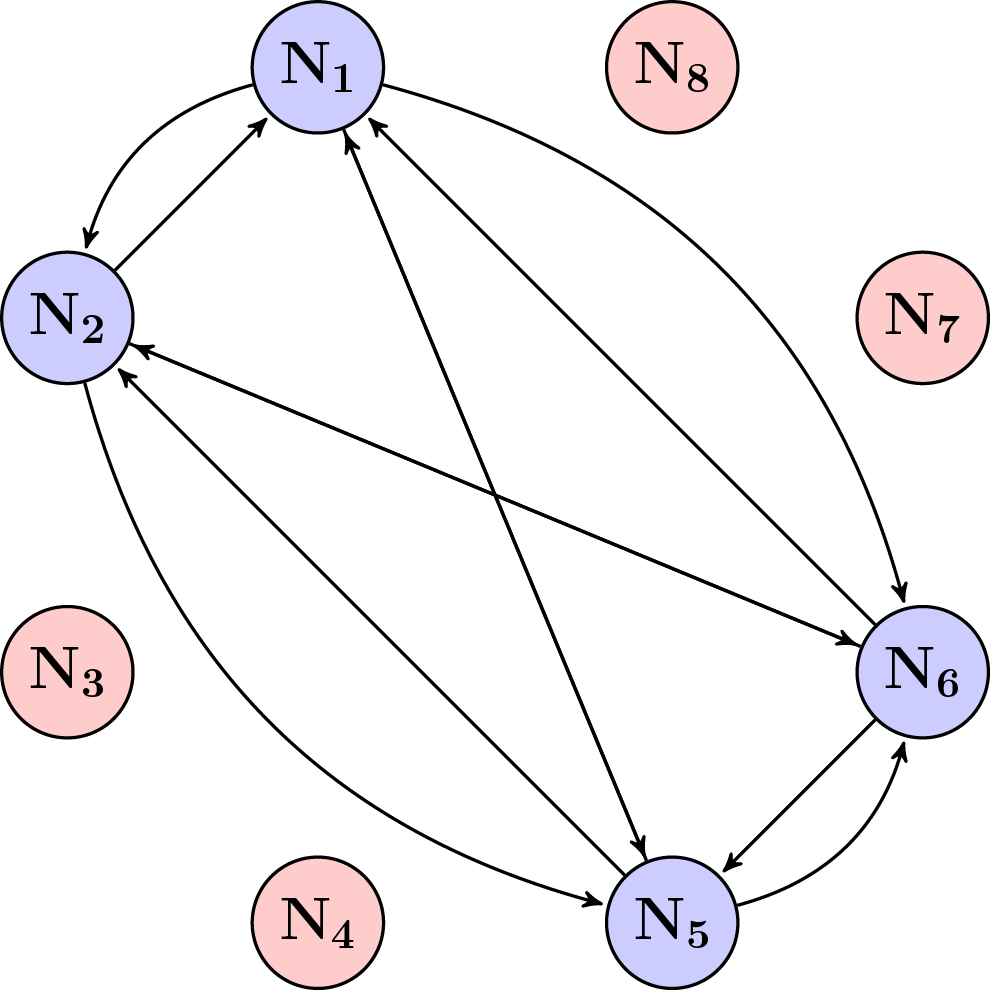}
\end{tabular}%
\caption{Fault-tolerance ability with loss of limb scenario. The limb 2 becomes inoperable, thus, not allowing transition of robots into states corresponding to nodes $\mathbf{N_3,N_4,N_7,N_8}$. The nodes are isolated and optimization can be performed to obtain desired control sequences without re-learning state transition rewards.}
\label{Fig:FaultTolerance}
\end{figure}

\begin{figure}[h]
\centering
\subfloat[]{
\includegraphics[scale=\thescale]{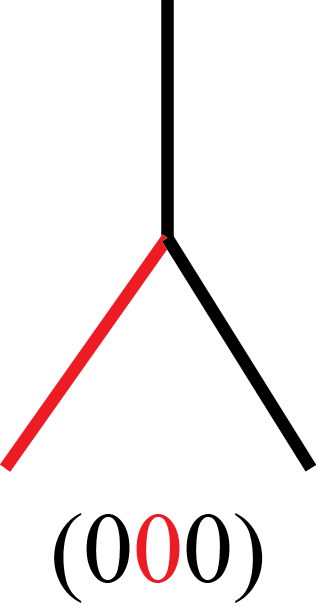} 
\includegraphics[scale=\thescale]{state-arrow-01.png}  
\includegraphics[scale=\thescale]{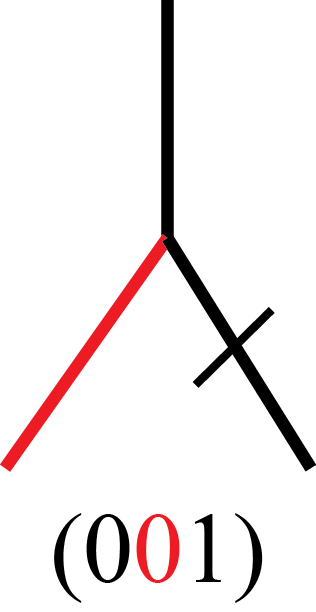} 
\includegraphics[scale=\thescale]{state-arrow-01.png}  
\includegraphics[scale=\thescale]{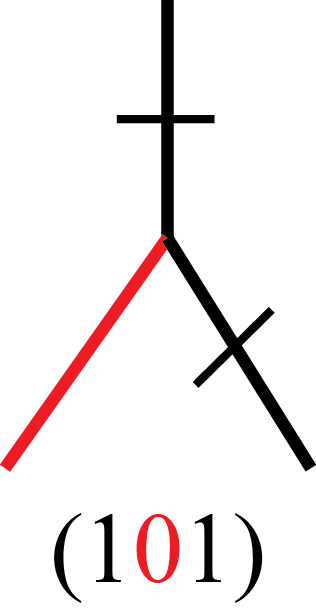} 
\includegraphics[scale=\thescale]{state-arrow-01.png}  
\includegraphics[scale=\thescale]{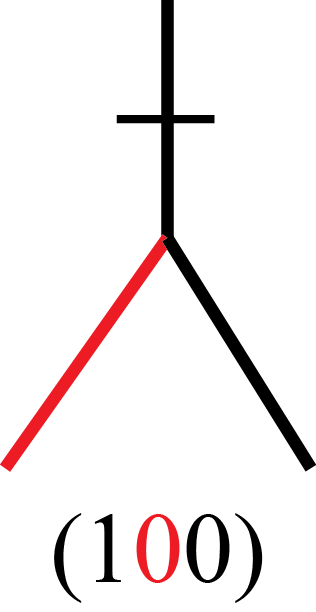} 
\includegraphics[scale=\thescale]{state-arrow-01.png}  
\includegraphics[scale=\thescale]{state-000-fault-01.png} 
}
\newline
\subfloat[]{
\includegraphics[scale=\thescale]{state-000-fault-01.png} 
\includegraphics[scale=\thescale]{state-arrow-01.png}  
\includegraphics[scale=\thescale]{state-001-fault-01.png} 
\includegraphics[scale=\thescale]{state-arrow-01.png}  
\includegraphics[scale=\thescale]{state-101-fault-01.png} 
\includegraphics[scale=\thescale]{state-arrow-01.png}  
\includegraphics[scale=\thescale]{state-000-fault-01.png} 
}
\caption{Two control sequences of same cost resulting from optimization of the modified graph for forward translation ($+X$ displacement). The red color corresponds to the inoperable limb.}
\label{Fig:FaultX+Translation}
\end{figure}

\begin{figure}[h]
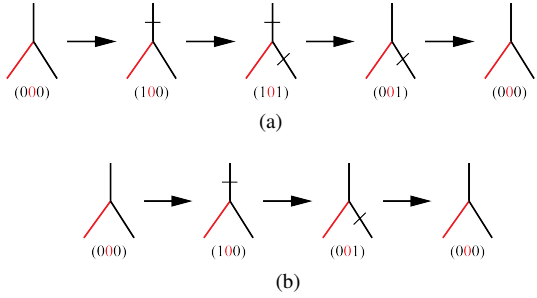

\centering
\subfloat[]{
\includegraphics[scale=\thescale]{state-000-fault-01.png} 
\includegraphics[scale=\thescale]{state-arrow-01.png}  
\includegraphics[scale=\thescale]{state-100-fault-01.png} 
\includegraphics[scale=\thescale]{state-arrow-01.png}  
\includegraphics[scale=\thescale]{state-101-fault-01.png} 
\includegraphics[scale=\thescale]{state-arrow-01.png}  
\includegraphics[scale=\thescale]{state-001-fault-01.png} 
\includegraphics[scale=\thescale]{state-arrow-01.png}  
\includegraphics[scale=\thescale]{state-000-fault-01.png} }
\newline
\subfloat[]{
\includegraphics[scale=\thescale]{state-000-fault-01.png} 
\includegraphics[scale=\thescale]{state-arrow-01.png}  
\includegraphics[scale=\thescale]{state-100-fault-01.png} 
\includegraphics[scale=\thescale]{state-arrow-01.png}  
\includegraphics[scale=\thescale]{state-001-fault-01.png} 
\includegraphics[scale=\thescale]{state-arrow-01.png}  
\includegraphics[scale=\thescale]{state-000-fault-01.png} 
}
\caption{Optimized state control sequences for backward translation for the faulty robot with second (red color) inoperable limb.}
\label{Fig:FaultX-Translation}
\end{figure}

\textit{Fault tolerance:} A loss of limb scenario is illustrated when the second actuator/sub-system of the three-limb robot becomes inoperable. Consequently, the robot cannot transition into or out of states $(010), (011), (110), (111)$ (Appendix \ref{App:Visualization}). The graph structure is modified by isolating the nodes corresponding to these states $\mathbf{N_3,N_4,N_7,N_8}$ as shown in Figure \ref{Fig:FaultTolerance}. The optimization can be applied to the modified graph to calculate desired control sequences without re-learning the state transition rewards. The optimized control sequences for this graph resulting in $+X$, $-X$ translation are illustrated in Figures \ref{Fig:FaultX+Translation} and \ref{Fig:FaultX-Translation} respectively.

The supplemental video illustrates the soft robot executing simple cycles to translate in forward and backward directions for both normal and limb loss scenario.

%
%
\section{Conclusion}
The research presents a data-driven, reinforcement learning inspired model-free control framework that indirectly models the robot-environment interaction and is summarized as a four-step process of discretization, visualization, learning and optimization. The dominant factors of robot-environment interaction are discretized into a finite number of behaviors. In this case, these behaviors correspond to one of two friction conditions and the combination of multiple limb behaviors define robot states. This discretization also allows the framework to be generic enough to be adaptable to a variety of different materials and actuator types. The framework utilizes graph theory language to describe control of soft robot. The finite number of robot states are represented by the nodes of the directed graph. Similarly, the transitions between states are represented by the directed arcs whereas the arc weights correspond to the result of the robot transitioning from one state to another. This state transition reward is dependent on the type of contact and needs to be learned for locomotion on different surfaces. This flexibility to learn the state transition rewards can facilitate adaptation to unexpected changes in the environment. The use of graph theory facilitates mathematical definition of periodic motions as simple cycles. These simple cycles allow formulation of an Integer Linear Programming problem that can be solved quickly using standard solvers. Furthermore, the graph representation imparts fault tolerance ability to the robot e.g. in case of a loss of limb scenario, graph nodes are isolated and new control sequences are calculated by performing optimization on the modified graph without needing to re-learn state transition rewards. %
A three limbed soft robot is controlled using the presented framework. The state transition rewards are visually recorded and multiple simple cycles are obtained for translation in forward and backward directions. %
This framework can be extended to produce more complex gaits, including highly nonlinear ones.
\section*{Acknowledgment}
Vishesh Vikas and Barry Trimmer are partly funded in part by the National Science Foundation grant IOS-1050908 to Barry Trimmer and National Science Foundation Award DBI-1126382 to Robert Peattie, Robert White and Barry Trimmer (Tufts University).

\appendices
\section{Visual representation of robot states} \label{App:Visualization}
The crossed marked limb indicates state $1$ or activated actuator ($\psi>\psi^*$), while the unmarked limb indicates state $0$ or relaxed actuator ($\psi<\psi^*$).
\begin{center}
\includegraphics[scale=\thescale]{state-000-01.png} \quad 
\includegraphics[scale=\thescale]{state-001-01.png} \quad
\includegraphics[scale=\thescale]{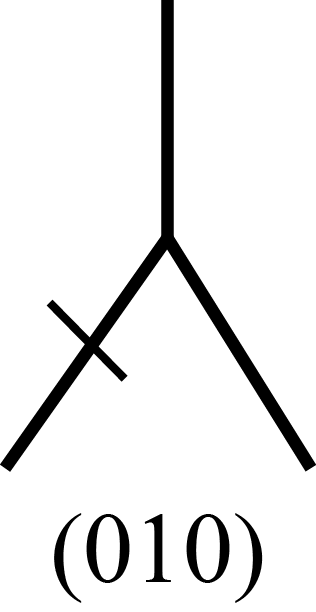} \quad
\includegraphics[scale=\thescale]{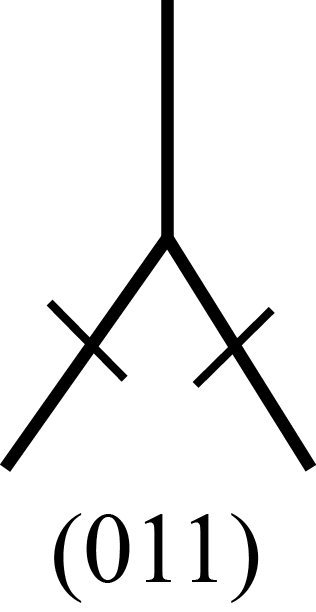} \quad
\includegraphics[scale=\thescale]{state-000-01.png} \quad 
\includegraphics[scale=\thescale]{state-001-01.png} \quad
\includegraphics[scale=\thescale]{state-010-01.png} \quad
\includegraphics[scale=\thescale]{state-011-01.png} 
\end{center}
\section{State transition rewards for three-limbed experimental robot} \label{App:Rewards}
\newcommand{\RewardRecord}[5]{{\small $#1 \rightarrow #2$} & {\small $[#3, #4, #5]$}}
\begin{tabular}{|c|c|c|c|}
\hline
{\scriptsize {\textsc{Transition}}} & {\scriptsize \textsc{Rewards}($R_i^T$)} &
{\scriptsize {\textsc{Transition}}} & {\scriptsize \textsc{Rewards}($R_i^T$)}\\
\hline
\RewardRecord{1}{2}{0}{0}{0} & \RewardRecord{5}{1}{0}{0}{0}\\
\RewardRecord{1}{3}{0}{0}{0} & \RewardRecord{5}{2}{-3}{-1}{-1}\\
\RewardRecord{1}{4}{0}{0}{0} & \RewardRecord{5}{3}{-2}{-1}{5}\\
\RewardRecord{1}{5}{0}{0}{0} & \RewardRecord{5}{4}{-3}{-1.5}{-15}\\
\RewardRecord{1}{6}{0}{0}{5} & \RewardRecord{5}{6}{-4}{-1}{-2}\\
\RewardRecord{1}{7}{0}{0}{-5} & \RewardRecord{5}{7}{-4}{0}{2}\\
\RewardRecord{1}{8}{1}{0.5}{0} & \RewardRecord{5}{8}{-1}{0}{0}\\
\hline
\RewardRecord{2}{1}{0}{0}{0} & \RewardRecord{6}{1}{0}{0}{0}\\
\RewardRecord{2}{3}{0}{0}{0} &\RewardRecord{6}{2}{-2}{-0.5}{2}\\
\RewardRecord{2}{4}{0}{-0.5}{-10} & \RewardRecord{6}{3}{-1}{0.5}{10}\\
\RewardRecord{2}{5}{1}{0.5}{-10} & \RewardRecord{6}{4}{-3}{-3}{-15}\\
\RewardRecord{2}{6}{1}{0}{-1} & \RewardRecord{6}{5}{0}{0}{0}\\
\RewardRecord{2}{7}{2}{1}{-2} & \RewardRecord{6}{7}{0.5}{0}{2}\\
\RewardRecord{2}{8}{3.5}{0.5}{0} & \RewardRecord{6}{8}{3}{-1.5}{0}\\
\hline
\hline
\RewardRecord{3}{1}{0}{0}{0} & \RewardRecord{7}{1}{0}{0}{0}\\
\RewardRecord{3}{2}{0}{0}{0} & \RewardRecord{7}{2}{-2}{0}{0}\\
\RewardRecord{3}{4}{0}{0.5}{15} & \RewardRecord{7}{3}{-2}{0.5}{-2}\\
\RewardRecord{3}{5}{2}{-0.5}{30} & \RewardRecord{7}{4}{-0.5}{0.25}{0}\\
\RewardRecord{3}{6}{1}{0}{10} & \RewardRecord{7}{5}{-0.5}{-0.25}{0}\\
\RewardRecord{3}{7}{1}{0}{1} & \RewardRecord{7}{6}{-1}{0}{0}\\
\RewardRecord{3}{8}{2}{-0.75}{7.5} & \RewardRecord{7}{8}{3}{1.5}{0}\\
\hline
\RewardRecord{4}{1}{0}{0}{0} & \RewardRecord{8}{1}{1}{0.5}{0}\\
\RewardRecord{4}{2}{0}{-0.5}{-10} & \RewardRecord{8}{2}{-2.5}{-2.5}{-4}\\
\RewardRecord{4}{3}{0}{-0.5}{-10} & \RewardRecord{8}{3}{0.5}{0.5}{2}\\
\RewardRecord{4}{5}{4}{0.5}{15} & \RewardRecord{8}{4}{-1}{-1}{0}\\
\RewardRecord{4}{6}{3}{0}{-6} & \RewardRecord{8}{5}{5}{0}{0}\\
\RewardRecord{4}{7}{2}{0.5}{0} & \RewardRecord{8}{6}{3}{-1.5}{0}\\
\RewardRecord{4}{8}{3}{-0.5}{0} & \RewardRecord{8}{7}{1}{0.5}{0}\\
\hline
\end{tabular}

\bibliographystyle{IEEEtr}
\bibliography{IEEEabrv,SoftRefs}

\begin{thebibliography}{10}

\bibitem{laschi_design_2009}
C.~Laschi, B.~Mazzolai, V.~Mattoli, M.~Cianchetti, and P.~Dario, ``Design of a
  biomimetic robotic octopus arm,'' {\em Bioinspiration \& Biomimetics},
  vol.~4, no.~1, p.~015006, 2009.

\bibitem{cianchetti_new_2009}
M.~Cianchetti, V.~Mattoli, B.~Mazzolai, C.~Laschi, and P.~Dario, ``A new design
  methodology of electrostrictive actuators for bio-inspired robotics,'' {\em
  Sensors and Actuators B: Chemical}, vol.~142, no.~1, pp.~288--297, 2009.

\bibitem{walker_continuum_2005}
I.~D. Walker, D.~M. Dawson, T.~Flash, F.~W. Grasso, R.~T. Hanlon, B.~Hochner,
  W.~M. Kier, C.~C. Pagano, C.~D. Rahn, and Q.~M. Zhang, ``Continuum robot arms
  inspired by cephalopods,'' vol.~5804, pp.~303--314, 2005.

\bibitem{hirose_biologically_2004}
S.~Hirose and M.~Mori, ``Biologically inspired snake-like robots,'' in {\em
  {IEEE} {International} {Conference} on {Robotics} and {Biomimetics}},
  pp.~1--7, 2004.

\bibitem{wright_design_2007}
C.~Wright, A.~Johnson, A.~Peck, Z.~McCord, A.~Naaktgeboren, P.~Gianfortoni,
  M.~Gonzalez-Rivero, R.~Hatton, and H.~Choset, ``Design of a modular snake
  robot,'' in {\em {IEEE}/{RSJ} {International} {Conference} on {Intelligent}
  {Robots} and {Systems}}, pp.~2609--2614, Oct. 2007.

\bibitem{shepherd_multigait_2011}
R.~F. Shepherd, F.~Ilievski, W.~Choi, S.~A. Morin, A.~A. Stokes, A.~D. Mazzeo,
  X.~Chen, M.~Wang, and G.~M. Whitesides, ``Multigait soft robot,'' {\em
  Proceedings of the National Academy of Sciences}, vol.~108, pp.~20400--20403,
  Dec. 2011.

\bibitem{lin_goqbot:_2011}
H.-T. Lin, G.~G. Leisk, and B.~Trimmer, ``{GoQBot}: a caterpillar-inspired
  soft-bodied rolling robot,'' {\em Bioinspiration \& Biomimetics}, vol.~6,
  p.~026007, June 2011.

\bibitem{umedachi_highly_2013}
T.~Umedachi, V.~Vikas, and B.~A. Trimmer, ``Highly {Deformable} 3-{D} {Printed}
  {Soft} {Robot} {Generating} {Inching} and {Crawling} {Locomotions} with
  {Variable} {Friction} {Legs},'' in {\em {IEEE}/{RSJ} {International}
  {Conference} on {Intelligent} {Robots} and {Systems}}, 2013.

\bibitem{wang_locomotion_2014}
W.~Wang, J.-Y. Lee, H.~Rodrigue, S.-H. Song, W.-S. Chu, and S.-H. Ahn,
  ``Locomotion of inchworm-inspired robot made of smart soft composite
  ({SSC}),'' {\em Bioinspiration \& Biomimetics}, vol.~9, p.~046006, Dec. 2014.

\bibitem{rigatos_model-based_2009}
G.~G. Rigatos, ``Model-based and model-free control of flexible-link robots:
  {A} comparison between representative methods,'' {\em Applied Mathematical
  Modelling}, vol.~33, pp.~3906--3925, Oct. 2009.

\bibitem{hannan_kinematics_2003}
M.~W. Hannan and I.~D. Walker, ``Kinematics and the {Implementation} of an
  {Elephant}'s {Trunk} {Manipulator} and {Other} {Continuum} {Style}
  {Robots},'' {\em Journal of Robotic Systems}, vol.~20, no.~2, pp.~45--63,
  2003.

\bibitem{webster_design_2010}
R.~J. Webster and B.~A. Jones, ``Design and {Kinematic} {Modeling} of
  {Constant} {Curvature} {Continuum} {Robots}: {A} {Review},'' {\em The
  International Journal of Robotics Research}, vol.~29, pp.~1661--1683, Nov.
  2010.

\bibitem{renda_3d_2012}
F.~Renda, M.~Cianchetti, M.~Giorelli, A.~Arienti, and C.~Laschi, ``A 3d
  steady-state model of a tendon-driven continuum soft manipulator inspired by
  the octopus arm,'' {\em Bioinspiration \& Biomimetics}, vol.~7, p.~025006,
  June 2012.

\bibitem{duriez_control_2013}
C.~Duriez, ``Control of elastic soft robots based on real-time finite element
  method,'' in {\em {IEEE} {International} {Conference} on {Robotics} and
  {Automation}}, pp.~3982--3987, May 2013.

\bibitem{chirikjian_kinematics_1995}
G.~Chirikjian and J.~Burdick, ``The kinematics of hyper-redundant robot
  locomotion,'' {\em IEEE Transactions on Robotics and Automation}, vol.~11,
  pp.~781--793, Dec. 1995.

\bibitem{saunders_modeling_2011}
F.~Saunders, B.~A. Trimmer, and J.~Rife, ``Modeling locomotion of a soft-bodied
  arthropod using inverse dynamics,'' {\em Bioinspiration \& Biomimetics},
  vol.~6, p.~016001, Mar. 2011.

\bibitem{sutton_introduction_1998}
R.~S. Sutton and A.~G. Barto, {\em Introduction to {Reinforcement} {Learning}}.
\newblock Cambridge, MA, USA: MIT Press, 1st~ed., 1998.

\bibitem{antoulas_approximation_2005}
A.~Antoulas, {\em Approximation of {Large}-{Scale} {Dynamical} {Systems}}.
\newblock Advances in {Design} and {Control}, Society for Industrial and
  Applied Mathematics, Jan. 2005.

\bibitem{radhakrishnan_locomotion:_1998}
V.~Radhakrishnan, ``Locomotion: {Dealing} with friction,'' {\em Proceedings of
  the National Academy of Sciences}, vol.~95, pp.~5448--5455, May 1998.

\bibitem{gorb_attachment_2001}
S.~Gorb, {\em Attachment devices of insect cuticle}.
\newblock Springer, 2001.

\bibitem{diestel_graph_2010}
R.~Diestel, {\em Graph {Theory}}.
\newblock New York: Springer, 4th~ed., Oct. 2010.

\bibitem{johnson_finding_1975}
D.~Johnson, ``Finding {All} the {Elementary} {Circuits} of a {Directed}
  {Graph},'' {\em SIAM Journal on Computing}, vol.~4, pp.~77--84, Mar. 1975.

\bibitem{hagberg_aric_networkx._????}
{Hagberg, Aric}, {Schult, Dan}, and {Swart, Pieter}, ``{NetworkX}. {High}
  productivity software for complex networks.''

\bibitem{lopez_cesar_optimization_????}
{Lopez, Cesar}, ``Optimization {Techniques} {Via} {The} {Optimization}
  {Toolbox},'' in {\em {MATLAB} {Optimization} {Techniques}}.

\bibitem{optimization_gurobi_and_others_gurobi_????}
{Optimization, Gurobi and others}, ``Gurobi optimizer reference manual,'' tech.
  rep.

\end{thebibliography}
\end{document}